\documentclass[runningheads]{llncs}

\usepackage{moreverb,url}

\usepackage[colorlinks,bookmarksopen,bookmarksnumbered,citecolor=black,urlcolor=black]{hyperref}

\usepackage[font=footnotesize,skip=-1pt,justification=centering]{caption}
\usepackage{amsmath}
\usepackage[bottom]{footmisc} 
\usepackage{booktabs} 
\usepackage{caption} 
\usepackage{subcaption} 
\usepackage{graphicx}
\usepackage{pgfplots}
\usepackage[all]{nowidow}
\usepackage[utf8]{inputenc}
\usepackage{tikz}
\usetikzlibrary{er,positioning,bayesnet}
\usepackage{multicol}
\usepackage{algpseudocode,algorithm,algorithmicx}
\usepackage[inline]{enumitem} 
\usepackage{lscape}
\usepackage{longtable}
\usepackage{todonotes}
\usepackage{blindtext}
\usepackage{hyperref}

\begin{document}

\title{Home and destination attachment: study of cultural integration on Twitter}

\titlerunning{Home and destination attachment for migrants on Twitter}
\author{Jisu Kim\inst{1} \and
Alina S\^irbu\inst{2} \and
Giulio Rossetti \inst{3} \and
Fosca Giannotti\inst{3} \and
Hillel Rapoport \inst{4}}

\authorrunning{J. Kim et al.}

\institute{Scuola Normale Superiore, Pisa, Italy.
\email{jisu.kim@sns.it}\\ \and
University of Pisa, Pisa, Itlay.
\email{alina.sirbu@unipi.it} \and
Istituto di Scienza e Tecnologie dell'Informazione, National Research Council of Italy, Pisa, Italy. \email{{fosca.giannotti, giulio.rossetti}@isti.cnr.it}\and
Paris School of Economics, Université Paris 1 Panthéon-Sorbonne, \& CEPII, Paris, France. \email{hillel.rapoport@psemail.eu}
}

\maketitle  

\begin{abstract}
The cultural integration of immigrants conditions their overall socio-economic integration as well as natives' attitudes towards globalisation in general and immigration in particular. At the same time, excessive integration -- or acculturation -- can be detrimental in that it implies forfeiting one's ties to the home country and eventually translates into a loss of diversity (from the viewpoint of host countries) and of global connections (from the viewpoint of both host and home countries).
Cultural integration can be described using two dimensions: the preservation of links to the home country and culture, which we call \emph{home attachment}, and the creation of new links together with the adoption of cultural traits from the new residence country, which we call \emph{destination attachment}. In this paper we introduce a means to quantify these two aspects based on Twitter data. 
We build home and destination attachment indexes and analyse their possible determinants (e.g., language proximity, distance between countries), also in relation to Hofstede's cultural dimension scores. The results stress the importance of host language proficiency to explain destination attachment, but also the link between language and home attachment. In particular, the common language between home and destination countries corresponds to increased home attachment, as does low proficiency in the host language. Common geographical borders also seem to increase both home and destination attachment. Regarding cultural dimensions, larger differences among home and destination country in terms of Individualism, Masculinity and Uncertainty appear to correspond to larger destination attachment and lower home attachment.

\end{abstract}

\newpage
\section{Introduction}

   The cultural integration of immigrants is a first-order social, political and economic issue. For the individual immigrant, it conditions his or her economic success and overall social integration to the host society. From the viewpoint of the latter, the promotion of the cultural integration of its immigrants has become a political imperative in these times of rising populism and cultural backlash against globalisation in general and immigration in particular (e.g., \cite{norris2019cultural}).\footnote{Norris, P. and Inglehart, R.F. (2019): Cultural backlash: Trump, Brexit, and Authoritarian Populism, Cambridge University Press).} However, from both the individual and social perspectives, too much cultural integration (or acculturation) may be detrimental: in terms of immigrants' subjective wellbeing, and in terms of lost diversity (from the viewpoint of host countries) and of global connections (from the viewpoint of both host and home countries). 
    In other words, it is in the best interests of all stakeholders to find the right balance between acculturation and cultural separatism, between loyalty to the home country and the host country cultures. Successful cultural integration brings new opportunities and, with them, an overall improvement of living conditions and well-being. Failure to integrate migrants in the host country's society may result in social conflict and cultural polarisation. 

   Cultural integration has been long studied by various research communities. These include international economic organisations  which have built indicators for integration at different levels, considering socio-economic features such as labour market participation, living conditions, civic engagement and social integration \cite{eurostat2011indicators,oecd2018settling,huddleston2013using}. 
    On the other hand, studies of integration have been mainly done by sociologists, by employing survey data such as World Values Survey, Eurobarometer, and European Social Survey.
    The main elements used in the studies are often inter-marriage, religion and language \cite{vigdor2008measuring,lochmann2019effect,sirbu2020human,esser2006migration}. 
    However, studying integration is very complex, as one is ``not only attracted to the culture of host society but is also held back from his culture of origin" \cite{park1928human,safi2008immigrant}.
    The four-fold model reflects this complexity by dividing acculturation into four different classes: \emph{assimilation}, \emph{integration}, \emph{marginalisation} and \emph{separation}.  
    \cite{constant2008measuring,council1997measurement,portes1985latin,penninx2003integration,berry1997immigration}.  
    Integration takes place when a migrant's and receiving society's characteristics mutually accommodate. Assimilation on the other hand takes place when a migrant perfectly absorbs the characteristics of the receiving society, losing the connection to the home country. Marginalisation refers to a situation where migrants remain distinguishable from the both of receiving and home society, whereas separation refers to complete rejection of host's culture. These theories typically consider two dimensions: preservation of links to the home country and cultural traits, which we call here \emph{home attachment, (HA)}, and formation of new links and adopting cultural traits from the country of migration, that we define as \emph{destination attachment (DA)}. Based on these two concepts, we can summarise the four integration patterns from the literature, as displayed in Table~\ref{tab:theories}.

    \begin{table}[b]
        \caption{Theories of integration and their relation to HA and DA.}
        \label{tab:theories}
        \centering
        \begin{tabular}{|l|l|l|}
        \hline
        & Low HA & High HA  \\ \hline
        Low DA & Marginalisation & Separation \\ \hline
        High DA &Assimilation & Integration \\ \hline
        \end{tabular}
    \end{table}
 
    In this paper we provide a novel method to compute HA and DA from Twitter data, to answer the following questions: \textit{How much do migrants absorb the culture of their destination society? Do they loose connection with their home country?} This is based on the topics that migrants and natives discuss on Twitter, through the analysis of hashtags. The HA index is defined as the fraction of tweets of a migrant that discuss topics related to their home country. Similarly, DA is the fraction of tweets discussing topics related to the destination country. These definitions are based on the idea that the topics discussed provide indications on various aspects of attachment: the amount of information that a person holds about a specific country, the social links to people living in a certain country, the interest in political and public issues of a country, adoption of customs and ideas, all related to integration as a wider concept.

    The analytic process that we introduce here includes three stages, and is based on a Twitter dataset containing data on users, their friends and their statuses. The first stage is to identify migrants by assigning a residence and nationality to Twitter users, starting from a previously developed method~\cite{kim2020digital}. The second stage is to determine country-specific topics by assigning nationalities to hashtags. The final stage is to compute the HA and DA indices for each migrant in our data. We examined the two indices in various settings,  to demonstrate their validity. First, we analysed the relationship between the two indices and compared them to a null model obtained by shuffling the hashtags in our dataset. Second, we studied different country-specific cases, i.e., immigrants in the United States and the United Kingdom, and emigrants from Italy. The indices were then compared with Hofstede's cultural dimension scores~\cite{hofstede1984culture} as well as other related variables such as distance and language proximity measures.

    The rest of the paper is organised as follows. In the next section we describe related work that studies integration and acculturation of migrants both in the sociology literature and in recent big data studies. In Section~\ref{sec:methods}, we define our methodology to compute the HA and DA indices, including data collection (Section~\ref{sec:data}), assigning nationality and residence to users (Section~\ref{sec:ident}), assigning nationality to hashtags (Section~\ref{sec:hash_nat}) and calculating the indices (Section~\ref{sec:ai_index}).
    In Section~\ref{sec:results}, we present our results, while Section~\ref{sec:discussion} concludes the paper.

\section{Related works}
\label{sec:literature}
    It has long been in the core interests of sociologists to study cultural identity and integration of migrants. 
    Using survey data, many have studied the complexity of migrants' conversion of cultural identity in the receiving societies. 
    Although a uniform definition of a culture does not exist, one way to define it is the following; ``the beliefs, values, social perspective, traditions, customs, and language shared within a group" \cite{robinson2019dynamics}. Taking the elements stated in the definition into account, studies have looked at language, role of media, inter-marriage and religion\footnote{\url{https://migrationdataportal.org/themes/migrant-integration}} to study whether a migrant is culturally integrated in the society \cite{vigdor2008measuring,kaasa2016new,kaasa2014regional}. 
    In particular, language plays an important role in various aspects of integration. It increases labour force participation of migrants and bring positive impacts on practical aspects of life, for example making friends in the class or talking to the teacher \cite{lochmann2019effect,alba2002only,sirbu2020human,esser2006migration}. In our work we also underline the relation between language proficiency and our DA index.

    In recent years, social big data has been employed to study integration of migrants \cite{guidotti,dubois2018studying,stewart2019rock}. Retail data including shopping behaviour in a large supermarket chain was used in \cite{guidotti} to measure the conversion of migrants' consumption behaviour towards that of natives. Through a data-driven approach, they identified 5 groups of migrants that show different trends towards adopting new consumption behaviours. 
    In \cite{dubois2018studying}, the authors used data collected from the Facebook Marketing API containing information on the country of origin, age, residence, spoken language and others, including the ``likes" of individual users. They quantified assimilation by introducing a score that serves as a proxy for migrants assimilating to local population's interests, using the ``likes" used by the Facebook users.
    Following the work in \cite{dubois2018studying},
    \cite{stewart2019rock} studied Mexican immigrants in the U.S and their cultural assimilation in terms of musical taste using Facebook data. They looked at the similarity of immigrants to the host population in terms of musical preferences, also looking at the interests of users. Furthermore, they extended their analysis to understand the differences in assimilation scores  between ethnicity and generations across different demographic groups. 
    In a more recent work, \cite{vieira2020using} looked at the diffusion of Brazilian cuisine around the world and estimated cultural distance between countries. They computed a so called interest entropy to measure how the interests are distributed around the world. They showed that the presence2 of Brazilian migrants explains, in part, the presence of interests in Brazilian cuisine in the host country. Other related factors were geographical proximity, and linguistic similarity, factors that also appear important in our study.

    In this paper, we also employ social big data for the analysis which allows us to overcome some of the limitations  of using survey data. For instance, it allows us to cover a wider population throughout broader geographical areas. However, different from Facebook data, Twitter data does not provide interests of individual users in the form of ``likes''. 
    We thus build our DA and HA indicators through hashtags as a proxy for their interests. In the process, we also employ the Shannon entropy, but in a different way from \cite{vieira2020using}: we use it to filter out hashtags that are not country-specific.
    Learning from the previous studies in Sociology, our analysis also takes into account HA (home attachment), which has not been as widely studied in the literature. In addition, many of the studies have been conducted from the host country's point of view towards their receiving migrants. Here, we also look at emigrants overseas, allowing the home country to better understand the allocation of their citizens abroad.

\section{The home and destination attachment indices}
\label{sec:methods}

    We propose to study home and destination attachment through the Twitter lens. We consider the topics discussed by migrants as a proxy to their interests, opinions and also to the amount of information about the context they live in, and define two indices: destination attachment (DA) and home attachment (HA). The methodology includes various stages: data collection, identifying migrant users by automatically assigning a nationality and residence label, identifying country-specific topics by assigning a nationality to Twitter hashtags, and finally the calculation of the indices.  

\subsection{Data}
\label{sec:data}
    Our data collection strategy originated from the methodology developed by \cite{kim2020digital}. The starting point is a Twitter dataset collected by the SoBigData.eu  Laboratory \cite{coletto2017perception}. We extracted from this dataset all the geo-located tweets posted from Italy from August to October 2015. This allowed us to obtain a set of 34,160 individual users that were in Italy in that period, which we call the first layer users. For these users, we downloaded the friends, resulting in 258,455 users that we denominate as second layer users. For all of these users, we have also gathered their 200 most recent tweets. Different from the work of \cite{kim2020digital}, we further extended the dataset to obtain a larger number of migrants by extracting also the friends of the second layer users (i.e. the third layer), and their 200 most recent Tweets. After this process, the total number of users grew to 59,476,205. Our dataset, therefore, consists of three layers:  the core first layer users, their friends (second layer users) and the friends of the friends (third layer users). Our analysis concentrates on a subset of these users for which we have information about their friends, resulting in a total of 200,354 users. These are users from the the first and second layers (some overlap was present among the two layers).

    \begin{figure}[t]
        \centering
        \includegraphics[width=0.45\textwidth]{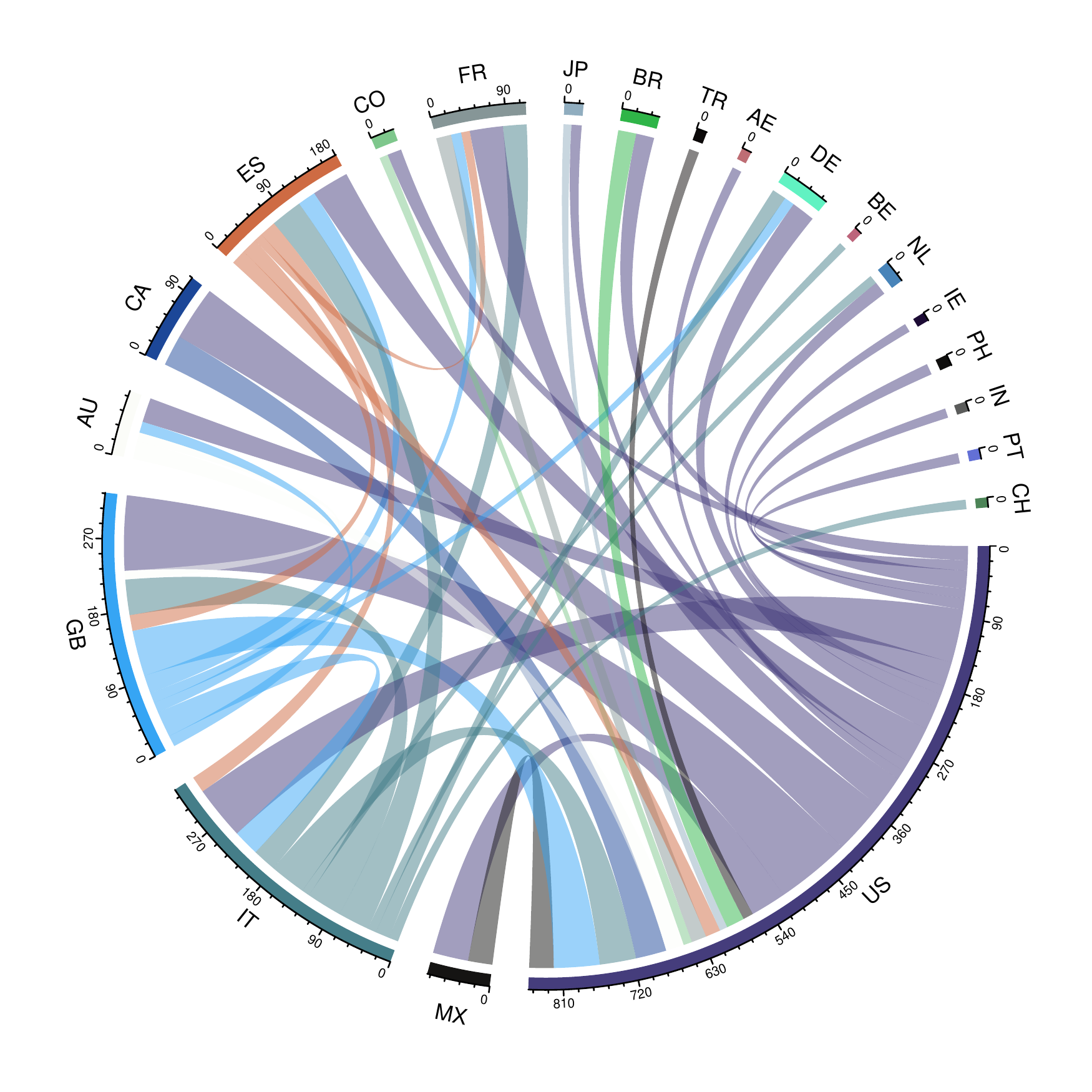}
        \caption{Chord diagram showing migration links between countries. The colour of the chord represents the nationality of the migrants, while the width of the chord represents the number of migrants in our dataset who had the 2018 residence in the corresponding destination country. For visualisation purposes  we show only 21 countries: those with at least 10 migrants.}
        \label{fig:chord}
    \end{figure}

 \subsection{Assigning residence and nationality to users}
 \label{sec:ident}
    In order to identify migrants in our dataset, we automatically assign to each user $u$ a nationality  country $C_n(u)$ and  a residence country $C_r(u)$ (for the year 2018) following the methodology in \cite{kim2020digital}. We define a migrant as ``a person who has the residence different from the nationality", i.e. $C_n(u) \neq C_r(u)$ .
    In order to identify a user's residence,
    we look at the number of days spent in each country in 2018 by looking at the time stamps and geo-locations of the tweets. The location where the user spent most of the time in 2018 is considered as the country of residence.
    On the other hand, the nationality is defined by looking at tweet locations of the user and user's friends. As shown in the study \cite{kim2020digital}, tweet language was not important in defining the nationality so we set the language weight to 0 here as well. 
    By comparing the country of residence and the nationality labels we were able to determine whether the user was a migrant or not in 2018. 
    
    Out of the total 200,354 users, we were able to identify nationalities of 197,464 users. As for the residence, we were able to identify residences of 57,299 users. 
    In total, we have identified both the residences and nationalities for 51,888 users. 
    Among 51,888 users, the total number of individuals users that we have identified as migrants are 4,940 users.
    We then filtered out users who have used less than 10 hashtags in 2018, leaving us with total of 3,226 migrant users.
    In Figure \ref{fig:chord}, we display the main migration links in our dataset: the number of migrants for countries that have at least 10 migrants, showing a total of 21 countries. However, overall, we have 128 countries of nationality and 163 countries of residence. From the plot, we see that in terms of nationality, the most present countries are the United States of America, Italy, Great Britain and Spain. This is due to the fact that our first level users were selected among those geo-localised in Italy. In terms of migration patterns, we note that Italy has mostly out-going links whereas countries like the USA and GB has a significant amount of both in and out-going links. France and Germany, on the other hand, have mostly in-coming links. 

\begin{figure}
        \centering
        \includegraphics[width=0.4\textwidth]{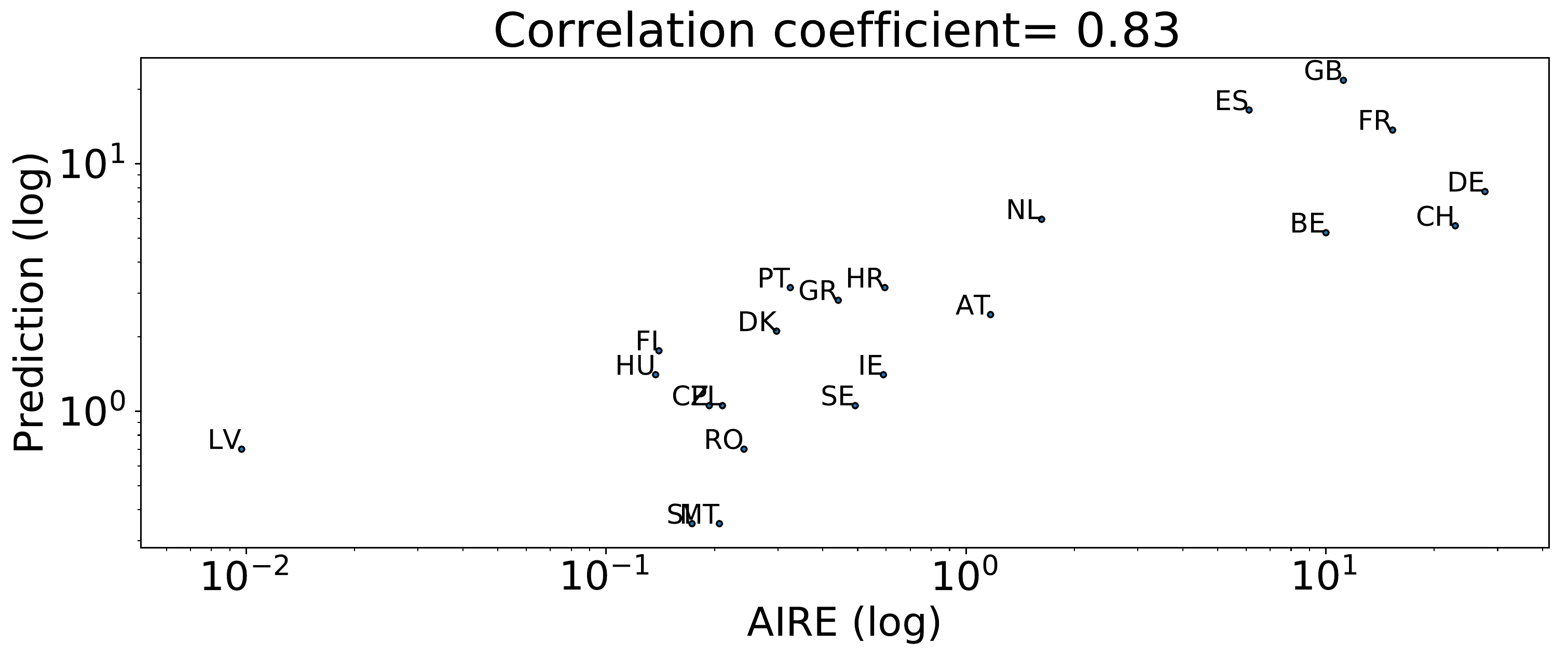}
        \includegraphics[width=0.4\textwidth]{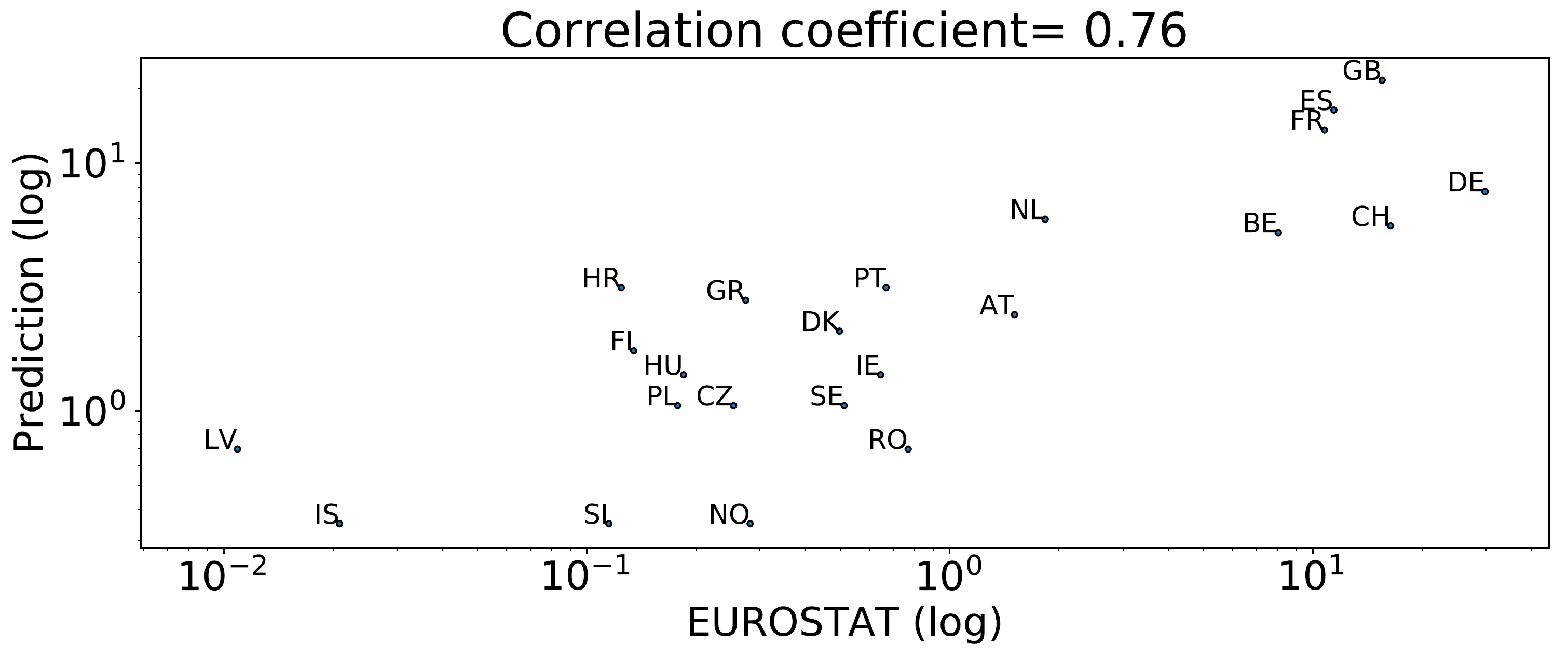}
        \includegraphics[width=0.4\textwidth]{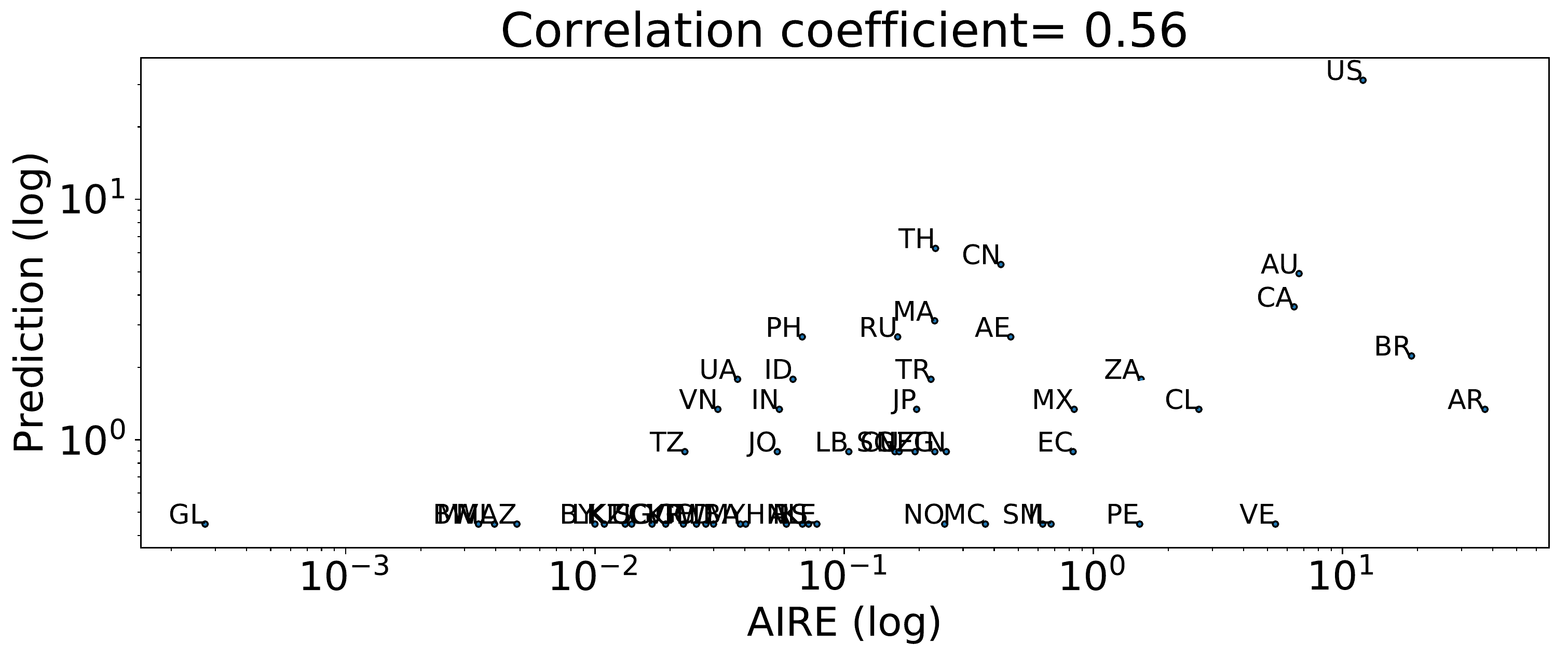}
        \caption{Percentage of Italian emigrants in various destination countries based on AIRE and Eurostat: predicted versus ground truth data.}
        \label{fig:ground_truth}
    \end{figure}

    We chose to employ this methodology because it adopts a definition of a migrant that is close to the official definition\footnote{Recommendations on Statistics of International Migration, Revision 1(p.113). United Nations, 1998, defines a migrant as “a person who moves to a country other than that of his or her usual residence for a period of at least a year”.}. It also allows us to identify both immigrants and emigrants simply by comparing the nationality and residence labels. It is important to mention that the migration patterns we see here are specific to our dataset, and are not meant to represent a global view of the world's migration. However we do observe some correlation to official data when looking at individual countries. In figure \ref{fig:ground_truth}, for instance, we show Spearman correlation coefficients between our predicted data and ground truth data for Italian emigrants from AIRE\footnote{Anagrafe degli italiani residenti all'estero (AIRE) is the Italian register data.} and Eurostat. For European countries, the correlation with the AIRE data is 0.831 and 0.762 with the Eurostat data. For non-European countries, the correlation stays at 0.56. This gives us reason to believe that this dataset can be used to validate our methodology of studying integration patterns through Twitter.

\subsection{Detecting country-specific topics}
\label{sec:hash_nat}

    The topics discussed on Twitter can be extracted through the analysis of hashtags. These are phrases that the users add to their tweets to mark the topic. In this analysis phase we detect country-specific topics by assigning nationalities to all the hashtags in our data.  To do this, for each hashtag we extract the list of users who use it, and we study the distribution of the nationality of all the users that are not labelled as migrants in the first stage (i.e. users who have the residence equal to the nationality). For those hashtags that appear mostly in one country (small entropy of the country distribution), we assign the nationality to the most frequent country. The hashtags that display a heterogeneous distribution across countries are not considered, since they are deemed international. 

    We begin by performing simple word processing for all the hashtags we have in the dataset. We selected all the hashtags used by non migrant users in 2018. We converted all the hashtags to lower case and removed signs such as comma, quotes, semicolons, and slashes. We removed also single characters. After the data cleaning process, we obtained a total of 639,494 hashtags that were used by non-migrants in 2018. 
    For each hashtag $h$, we define a dictionary where we store $P_h$, the distribution of the nationalities of the users using  hashtag $h$. Hence $P_h$ is a vector where for each country $c$ we have $P_h(c)$, the fraction, among all non-migrant users that use hashtag $h$, of users with nationality $c$. 
    Provided with this probability distribution, we compute the normalised entropy for each hashtag following Equation~\ref{eq:entropy}, where $\lvert P_h(c) \rvert$ is the cardinality of the dictionary $ P_h(c)$, i.e. the number  of countries where the hashtag is used. 
 
    \begin{equation}
    \label{eq:entropy}
        H(h)= \frac{-\sum_{c}P_h(c)\log P_h(c)}{\log(\lvert P_h(c) \rvert)}
    \end{equation}{}
    
\begin{figure}[t]
        \centering
        \includegraphics[width=0.4\textwidth]{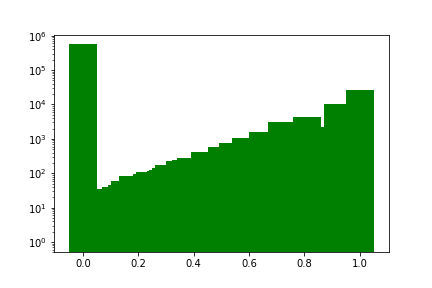}
        \caption{Entropy distribution in Log scale}
        \label{fig:entropy}
    \end{figure}

    Figure \ref{fig:entropy} displays the distribution of normalised entropy values across all hashtags in our dataset. We note that a majority of hashtags have zero entropy, hence they are mentioned in one country only, while a few show very high entropy levels, indicating they are international topics.

    To filter out international topics we select a threshold for the normalised entropy, that we here fix at the value 0.5.
   
    \begin{figure}[b]
        \centering
        \includegraphics[width=\textwidth]{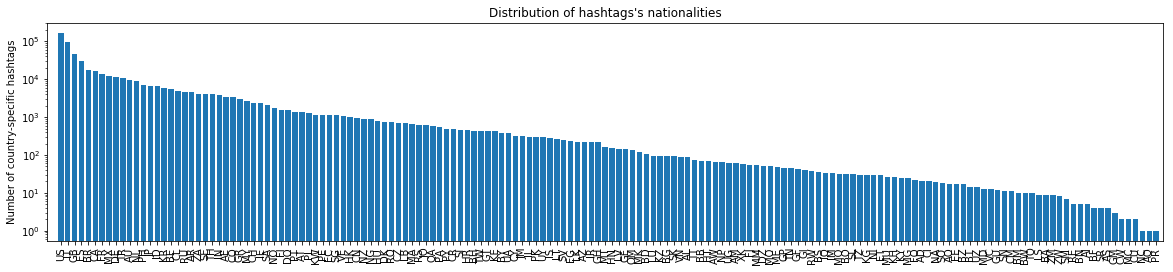}
        \caption{Distribution of hashtags' nationalities}
        \label{fig:dist_hashtag}
    \end{figure}

\subsection{Computing the home and destination attachment indices}
\label{sec:ai_index}
    Provided with the nationality of hashtags, we can define for each 3,226 migrant user the home and destination attachment, HA and DA. 
    Consider user $u$ with the country of nationality denoted as $C_n(u)$ and country of residence denoted as $C_r(u)$. To define the home attachment of user $u$, $HA(u)$, we consider $HT(u,C_n(u))$ the number of hashtags used by user $u$ specific to their country of origin, divided by $HT(u)$ the total number of hashtags of user $u$. For example, for an Italian national living in Korea, what fraction of their hashtags is Italian?

    \begin{equation}
        HA(u)= \frac{\#~ C_n(u)~ hashtags}{\# ~total ~ hashtags}=\frac{HT(u,C_n(u))}{HT(u)}
        \label{eq:attach}
    \end{equation}{}

    Similarly, the destination attachment index $DA$ is the fraction of hashtags they use that are labelled with their country of residence:
    
    \begin{equation}
    DA(u)= \frac{\#~ C_r(u)~ hashtags}{\# ~total ~ hashtags}=\frac{HT(u,C_r(u))}{HT(u)}
    \label{eq:integ}
    \end{equation}{} 
    Following the previous example, what is the fraction of Korean specific hashtags that the Italian emigrant is using?

    Both indices vary from 0 to 1. If they are equal to 1, it means that a migrant is either fully attached to the destination country or fully attached to home country. In contrast, indexes equal to 0 means that a migrant is either not attached to the destination country or not attached to the home country.  The sum of the two indices is always $\leq$ 1: a user cannot be fully attached to both home and destination, but has to `divide' their attention among the various countries they are interested in.

\section{Results}
\label{sec:results}

    \subsection{Overall distribution of DA and HA values}

 \begin{figure}[t]
        \centering
        \includegraphics[width=0.45\textwidth]{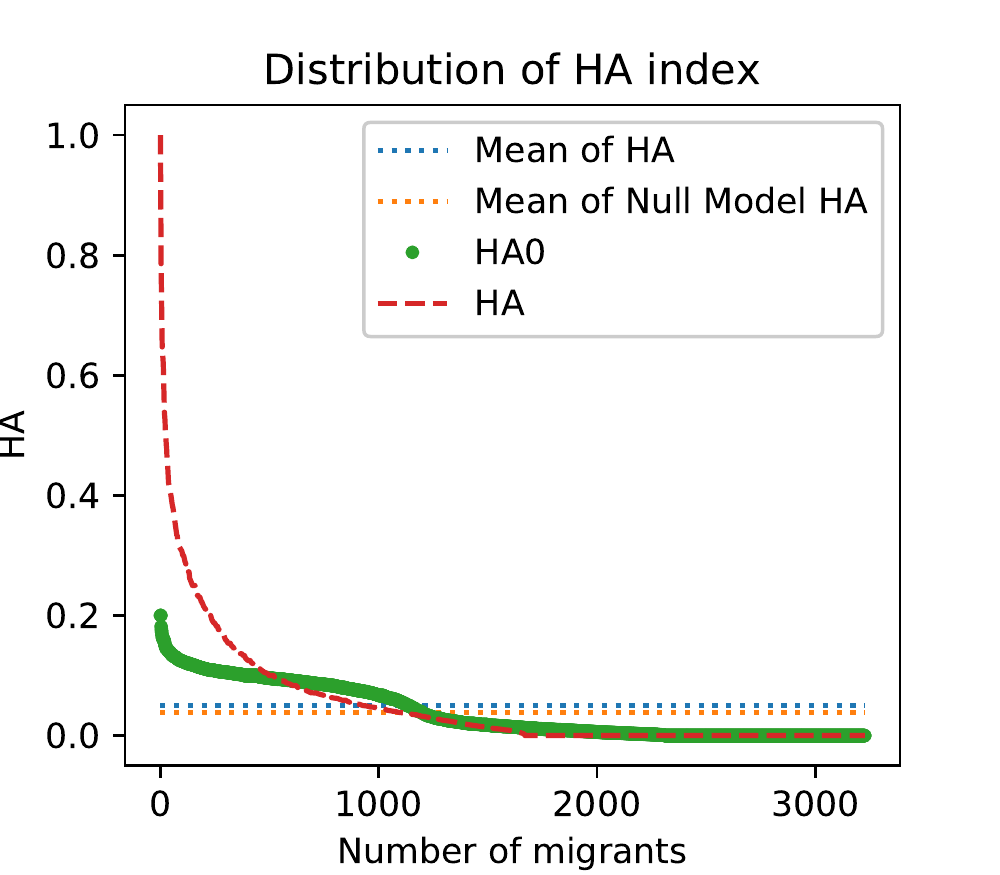}
        \includegraphics[width=0.45\textwidth]{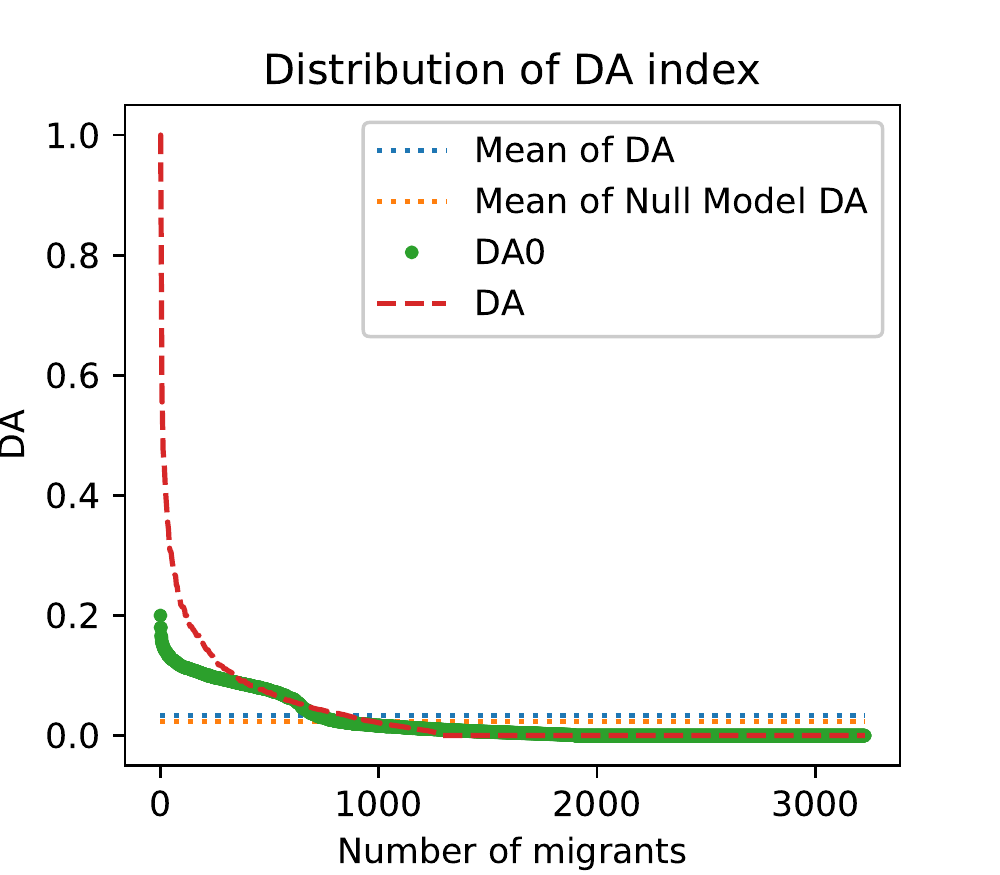}
        \caption{Distribution of HA and DA values, and comparison to null model HA$_0$ and DA$_0$. Means values are: $\bar{HA_0}= 0.038$ and $\bar{DA_0}=0.024$, $\bar{HA}=0.051$ and $\bar{DA}=0.034$.}
        \label{fig:shuffle2}

    \end{figure}

    The distributions of the home and destination attachment indices are shown in Figure~\ref{fig:shuffle2}. The HA index is 0.051 on average and the DA index is 0.034 on average for all the migrants we have in our dataset regardless of the nationality or the place of residence. We observe that some users have relatively high values for the two indices, however the majority are under 0.2 in both cases. 
    In the same figure, we compare these values with a null model analysis where the hashtags of individual users were randomly re-distributed five times. The null model tells us what the DA and HA values would be if users chose their topics of discussion randomly, i.e. there was no influence from the country of residence or nationality. We observe that in general the null model $DA_0$ and $HA_0$ are smaller than the actual index values, with lower means for the null model distributions.

    To statistically validate the difference between the null model, and DA and HA, we also computed two non-parametric tests: Wilcoxon and
    Kolmogorov– Smirnov (KS) tests. The results for the Wilcoxon test show that for both the DA and HA, their distributions are significantly different from the distribution of the $DA_0$ and $HA_0$ with p-values of 5.16$e^{-07}$ and 0.014, respectively.
    We obtained similar results from the KS tests, with p-values of 1.18$e^{-51}$ for DA and 2.98$e^{-56}$ for HA.
    Although not reported here, the results for KS-tests for sub-populations split by country of residence and country of origin equally show that the null model and the actual index values have different distributions.

\begin{figure}[t]

        \begin{center}
        \includegraphics[width=0.6\textwidth]{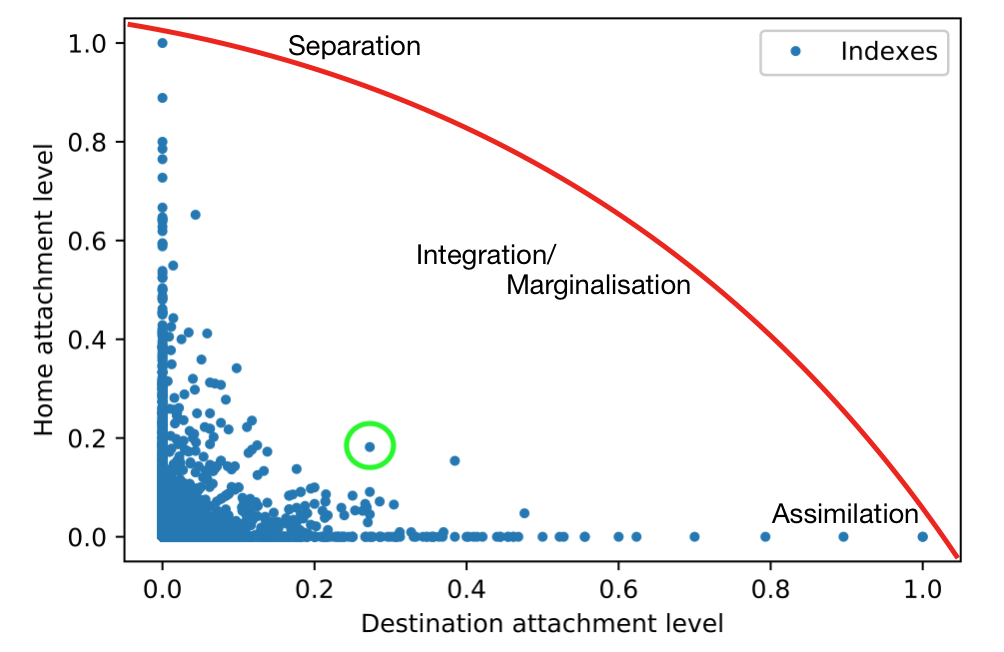}
        \caption{Pearson correlation between home and destination attachment indexes for all the migrants in the data: correlation coefficient: -0.13, p$-$value: $6.937e^{-14}$.}
        \label{fig:scatter}
        \end{center}

    \end{figure}

    To understand the relationship between the DA and HA, we computed the Pearson correlation among them. Figure~\ref{fig:scatter} displays the HA versus DA values for all users. A weak negative relation is found with $r=-0.13$, and p-value$=6.937e^{-14}$, indicating that in general the more a migrant is attached to his country of origin, the less the migrant is attached to the host country and vice versa. However, we can observe various different patterns for individual users, leading to different acculturation types as mentioned in Table~\ref{tab:theories}. In the same figure, the red curve provides an approximate indication of users' acculturation type. We underline the fact that we do not aim to provide a specific categorisation of acculturation types in this paper. Instead, we aim to provide a broad picture where the angle of each individual from the x/y-axis gives us an indication of the acculturation type. Thus, a migrant close to the x-axis is most probably going through and assimilation process,  a migrant close to the y-axis is undergoing separation, while those in between are undergoing integration or marginalisation. The distinction between integration and marginalisation depends on the length of the distance of data point from the origin. In other words, marginalisation is when the data point is close to 0 and integration is when the data point is point further away from 0. The data point circled in green would be a good example of an integrated migrant, who keeps good links with both home and destination country.

    \subsection{Language as a key factor for integration}

\begin{figure}

    \centering
        \includegraphics[width=0.6\textwidth]{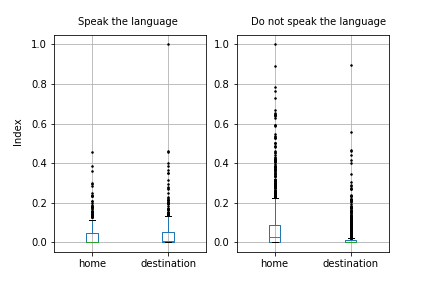}
        \caption{Box plots showing the HA and DA distributions for a group of migrants who speak the language of the host country on the left and a group of migrants who do not speak the language of the host country on the right. Means are $\bar{HA}=0.034$ and $\bar{DA}=0.041$ for users who speak the destination language, and $\bar{HA}=0.072$ and $\bar{DA}=0.019$ for those who do not speak it.}

    \label{fig:lang_att_integ}
    \end{figure}

    One possible candidate factor to explain the DA and HA values observed is language. As previously studied, language is considered to be a key factor in integration and our indexes reflect this importance as well. 
    In Figure~\ref{fig:lang_att_integ} we display the distribution of the DA and HA for two user groups: a group that speaks the language of the host country (i.e. over 90\% of their tweets are in that language) and a group that very rarely speaks the language of the host country (under 10\% of their tweets are in that language). Here, we are looking at all the migrants we have in the dataset regardless of the country of origin or the country of residence. We observe that the group that speaks the language of the destination country shows in general higher DA compared to the non-speaking group, confirming the significance of the language for integration in the host country. In addition, we observe that users who do not speak the language of the destination country tend to be more attached to their home country compared to those speaking the destination language. Hence, interestingly, destination language proficiency seems to affect both destination and home attachment levels. When comparing DA and HA within groups, the groups that speak the destination language have the two indices comparable, while for those who do not speak it, HA is much larger than DA, indicating a pattern of separation. However, we do not mean to generalise, what we observe are population level patterns. When looking at individual level, we do observe all four acculturation types discussed in Table~\ref{tab:theories}.

     {}

\subsection{Country-specific results}\label{sec:countries}

    In this section, we provide country-specific results. One of the advantage of using our methodology is that we can look at different countries simply by changing the labels. Hence, here we look at different country cases to understand how immigrants in a specific country behave and to know how emigrants from a certain country of origin behave in different countries. We selected three study cases which had the largest number of users in our data: immigrants in the US and UK, and emigrants from Italy. Here we consider only the migrant groups with at least 10 users. The square brackets in the figures below show the number of users we have for each country of origin.
    
    \subsubsection*{Immigrants in the United States}   
        \begin{figure}
        \centering
        \includegraphics[width=0.65\textwidth]{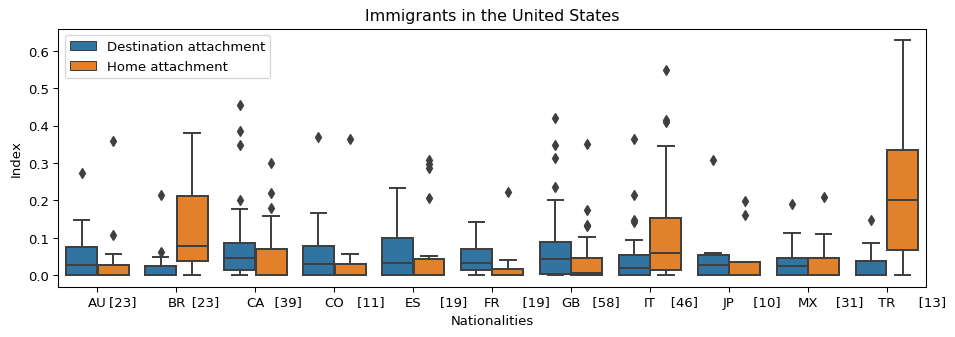}
        \includegraphics[width=0.34\textwidth]{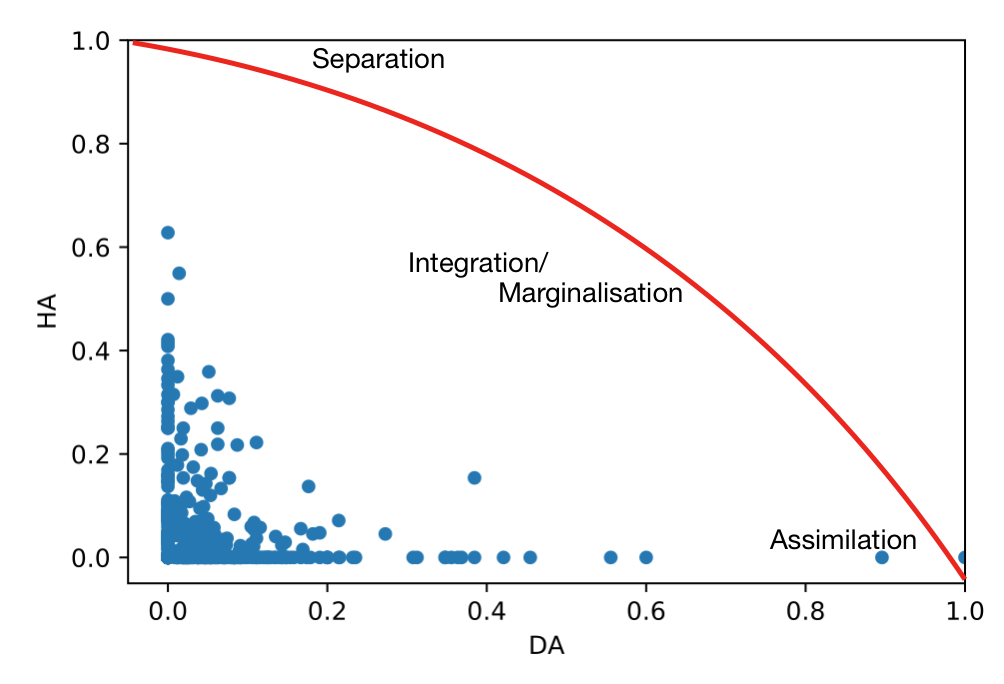}
        \caption{Left: Box plots for the DA and HA index of immigrants in the United States. Right: Scatter plot of HA vs. DA indicating approximate integration types for immigrants in the US.}
        \label{fig:mig_US}
        \end{figure}

    In Figure~\ref{fig:mig_US} on the left, we observe different destination and home attachment indices of 17 groups of immigrants from different countries of origin.
    Overall, we observe that for many groups of immigrants in the United States DA is larger than HA.
    Immigrants from Canada have the highest DA followed by Colombian and English immigrants. On the other hand, immigrants from Turkey have the highest HA followed by Brazilian and Italian immigrants. In the right figure, we observe data points individually on a scatter plot of HA vs. DA. It tells us that immigrants in the US are integrated and assimilated in general.

    We also compared our indexes to the work of Vigdor~\cite{vigdor2008measuring} that measures the degree of similarity between foreign-borns from different countries and natives in the United States. They measure three factors of assimilation: economic, cultural, civic, and their combination. The economic factor looks at employment status, income, education attainment and home ownership.
    The cultural factor looks at intermarriage, the ability to speak English, number of children and marital status. The civic factor looks at military service and citizenship. The composite factor is the overall score of the all three factors. Table~\ref{tab:vigdor} shows the Spearman correlation between our indices and the four factors of assimilation, trying to understand whether the attachment levels we see for each individual are similar to the assimilation levels Vigdor~\cite{vigdor2008measuring} found for nationals from the same countries. 
    The table shows that our DA and HA are most correlated with the cultural factor, followed by the economic factor. 
    It is interesting to remark that DA is positively correlated whereas HA is negatively correlated with the cultural factor of assimilation. 
    This tells us that for those nationalities for which Vigdor observed high cultural assimilation, we observe high DA and low HA, which is exactly how we propose to use our indices to describe assimilation (see Table~\ref{tab:theories} above).  
    A similar relation can be seen with the economic factor: nationalities with high economic assimilation levels also show high DA and low HA.
    Interestingly, the civic factor does not show the same relation: foreign-borns of nationalities that appear to be well assimilated from the civic point of view in Vigdor's work tend to show a high HA in our work, and no relation with DA. It appears thus that civic assimilation in the destination country corresponds also to a tighter relation with the home country of a migrant. 
    
    A caveat in looking at this table is that here we are looking at identified migrants and hashtags in 2018 and comparing them to the assimilation scores of 2006. There could be possible changes in immigrants' behaviours between 2006 and 2018. 
    A second caveat is that we are computing correlations at individual level, while Vigdor's scores are based on groups of migrants. Since there is variability among individuals, it is likely the case that two US immigrants with the same nationality will have different DA and HA scores in our data, while the Vigdor data will contain an unique score for them. This inevitably decreases correlations.

    \begin{table}[t]
    \centering
    
    \begin{tabular}{|l|l|l|l|l|l|l|}
    \hline
    & DA & HA & Composite & Economic & Cultural & Civic \\ \hline
    DA & 1.0*** & -0.231*** & 0.087 & 0.185*** & 0.198*** & 0.045 \\ \hline
    HA & -0.231*** & 1.0*** & 0.129** & -0.145** & -0.2*** & 0.159*** \\ \hline
    Composite & 0.087 & 0.129** & 1.0*** & 0.628*** & 0.406*** & 0.916*** \\ \hline
    Economic & 0.185*** & -0.145** & 0.628*** & 1.0*** & 0.766*** & 0.551*** \\ \hline
    Cultural & 0.198*** & -0.2*** & 0.406*** & 0.766*** & 1.0*** & 0.218*** \\ \hline
    Civic & 0.045 & 0.159*** & 0.916*** & 0.551*** & 0.218*** & 1.0*** \\ \hline
    \end{tabular}

    \caption{Spearman correlation table for immigrants in the United States:   Vigdor's assimilation scores and DA \& HA indices. Significance levels are marked with *** p-value $<$0.01, ** p-value $<$0.05, * p-value $<$0.1.}
    \label{tab:vigdor}
    \end{table}
    
    \subsubsection*{Immigrants in the United Kingdom}

\begin{figure}
        \centering
        \includegraphics[width=0.55\textwidth]{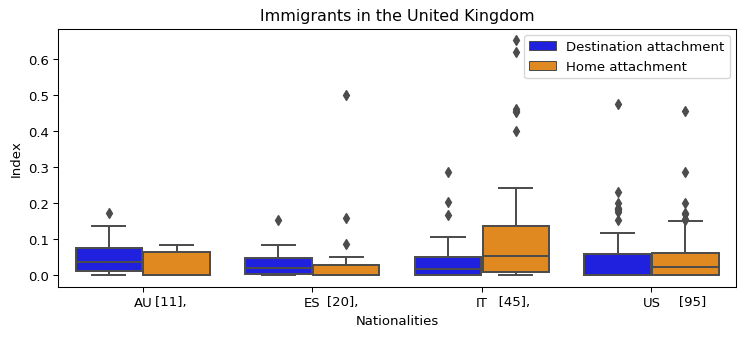}
        \includegraphics[width=0.4\textwidth]{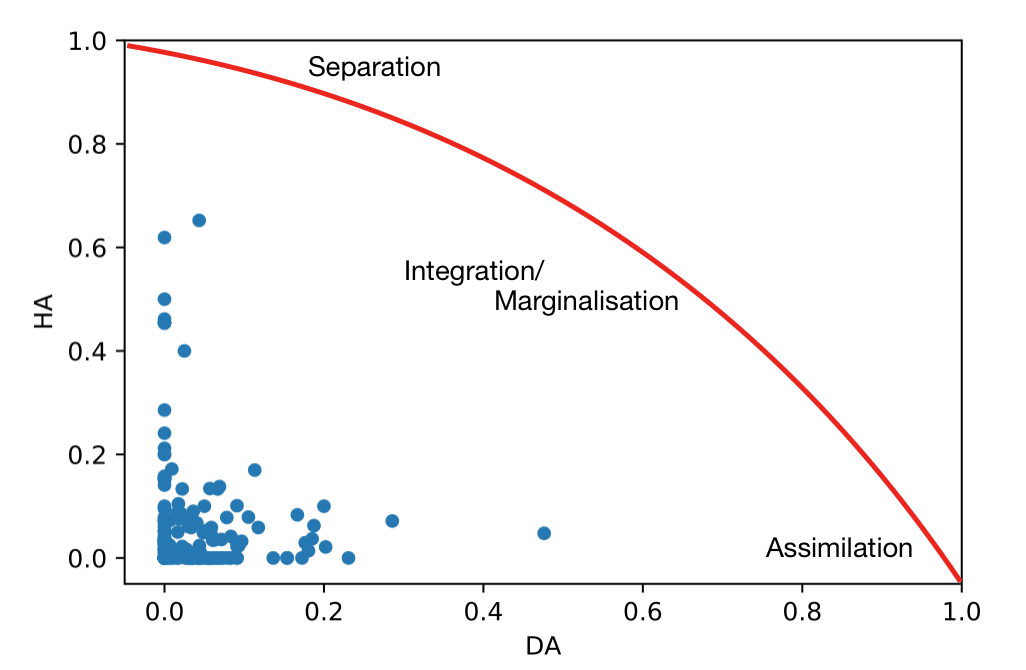}
        \caption{Left: Box plots for DA and HA for immigrants residing in the United Kingdom. Right: Scatter plot of HA vs. DA indicating approximate integration types for immigrants in the UK.}
        \label{fig:mig_gb}
    \end{figure}

    Figure~\ref{fig:mig_gb} shows the indices for the immigrants residing in the United Kingdom. Only four groups are shown, corresponding to those that have at least 10 migrants. Overall, UK immigrants in our data are more attached to home than to the destination country. On average, the DA is 0.04 and the HA is 0.063. 
    From the figure on the left, it is clear that immigrants from Italy have the highest HA index. On the other hand, we observe that immigrants from Australia that share long historical ties with the UK have the highest DA index.
    Looking at the figure on the right, we can observe that immigrants are mostly in the area of marginalisation/integration.

    \subsubsection*{Italian emigrants}

\begin{figure}
        \begin{center}
        \includegraphics[width=0.6\textwidth]{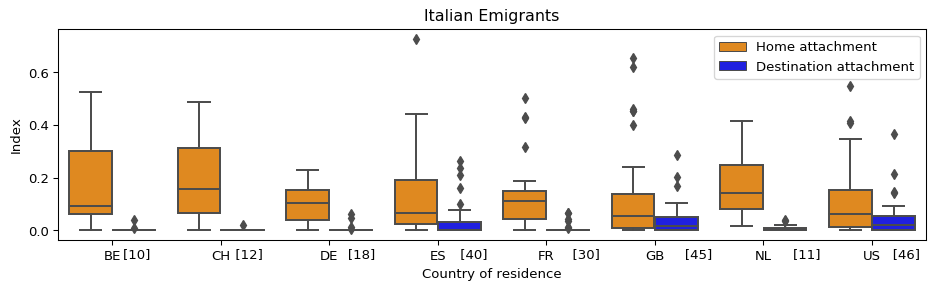}
        \includegraphics[width=0.35\textwidth]{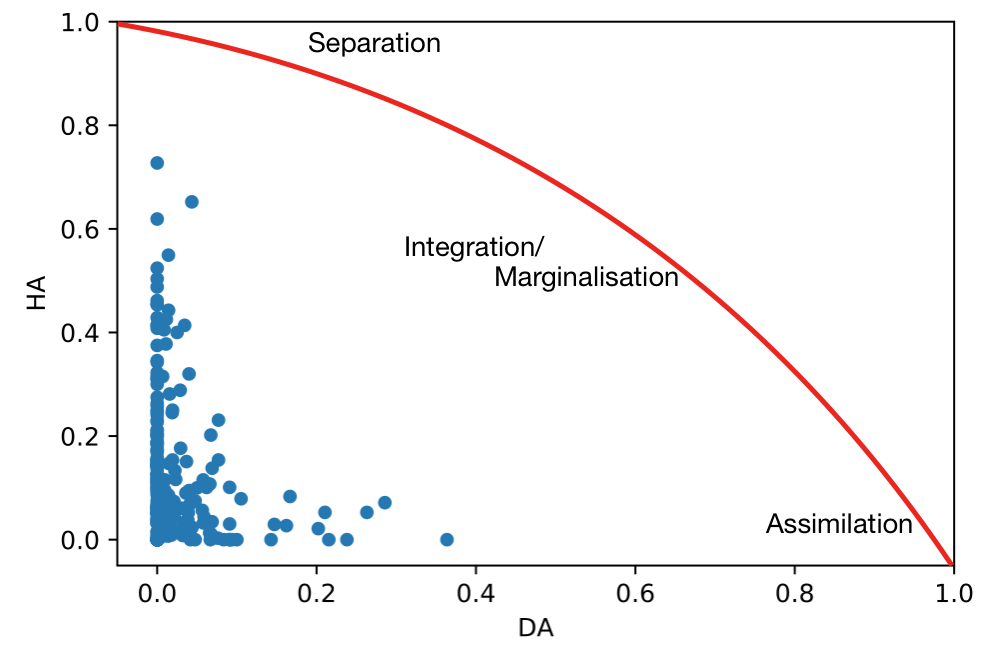}
        \caption{Left: Box plots for DA and HA  for Italian nationals living abroad. Countries on x-axis are countries of residence of Italians. Bottom: Scatter plot of HA vs. DA indicating approximate integration types for Italian emigrants.}
        \label{fig:box_ai_it}
        \end{center}
    \end{figure}

    Figure~\ref{fig:box_ai_it} displays the DA and HA indices for Italian emigrants across different countries of residence. 
    In general, we observe that Italians are more attached to their home country than to their destination country.
    Switzerland, Belgium and Netherlands are the three countries where Italian emigrants are most attached to home. On the other hand, Italians  tend  to show higher DA levels  in  English  speaking  countries: the US and in the UK. Among the higher DA levels we also observe Spain, probably due to the language similarity. In the figure on the right, we also observe that Italian emigrants have higher HA level compared to DA level. This data points indicate that they are in general close to the \textit{separation} type of acculturation.
    
    \subsection{Hofstede's cultural dimension scores and other measures}

    To further validate our indices, we have also compared our results with Hofstede's six cultural dimensions, plus various other language proximity measures and geographical distances \cite{hofstede1984culture,hofstede2011dimensionalizing,mayer2011notes,melitz2014native}.
    Hofstede's cultural dimensions are well known measures of culture, initially studied to better design the organisational context of business \cite{hofstede1984culture}. 
    According to his initial studies,
    cultures can be studied along four dimensions: power, masculinity, individualism, and uncertainty avoidance\footnote{Power distance: whether a hierarchical order is accepted among people. Masculinity (vs. Femininity): whether the country is driven by competition, achievement and success. Individualism (vs. Collectivism): how ``me-centred" the people are in the country. Uncertainty avoidance: how comfortable people are when faced with uncomfortable and ambiguous situations.}. 
    In his later studies, long-term orientation and indulgence\footnote{Long-term (vs. Short term) orientation: whether the importance is given to what has been done already or to the future, Indulgence (vs. restraint): how strict the people are towards their desires.} were added to the cultural dimensions \cite{hofstede2011dimensionalizing}. 
    To compare our indices with Hofstede's cultural dimensions, we computed the differences of scores between the home and the destination countries of migrants, as a measure of the cultural distance among countries. We then computed the correlation between our HA and DA indices and the cultural distances obtained. Hofstede's data contain a total of 114 countries, while our nationalities and residences cover 128 and 163 countries, respectively. Therefore we considered only users for which both nationality and residence were among the 114 countries, resulting in 3,082 users. 
    In addition to Hofstede's scores, we also added the following variables: distance between the capitals of the countries (\textit{distcap}), common native language (\textit{cnl}), common spoken language (\textit{csl}), and two dummy variables on whether the countries are sharing borders (\textit{contig}) and common official language (\textit{comlang$\_$off}). The \textit{cnl} and \textit{csl} variables vary at a scale between 0 to 1, indicating 0 if there are no commonality and 1 if they share full commonality.

    \begin{table*}[t]
    \centering
    \resizebox{\textwidth}{!}{
    \begin{tabular}{|l|l|l|l|l|l|l|l|l|l|l|l|l|l|}
    \hline
     & DA & HA & Power & Individualism & Masculinity & Uncertainty & Orientation & Indulgence & contig & comlang\_off & distcap & csl & cnl \\ \hline
    DA & 1.0*** & -0.153*** & -0.054*** & 0.155*** & 0.133*** & -0.046*** & -0.041** & 0.016 & 0.003 & 0.069*** & 0.034* & 0.083*** & 0.099*** \\ \hline
    HA & -0.153*** & 1.0*** & 0.029 & -0.092*** & -0.113*** & -0.014 & 0.026 & 0.03* & 0.063*** & -0.012 & -0.074*** & 0.023 & 0.021 \\ \hline
    \end{tabular}%
    }
    \caption{Correlation table for HA \& DA and Hofstede's cultural dimension scores for migrants at individual level. Significance levels are marked with *** p-value $<$0.01, ** p-value $<$0.05, * p-value $<$0.1.}
    \label{tab:hof_mig}
    \end{table*}

    Table~\ref{tab:hof_mig} shows the Pearson correlations computed at individual level. The first interesting remark is that in general our DA and HA indices behave differently across the six cultural dimensions, language and distance variables. This means that, when correlations are significant, when HA shows a positive relation,  DA shows a negative one and vice-versa. This is compatible with the fact that HA and DA are negatively correlated among themselves, meaning that, in general, as migrants becomes more attached to the destination they lose links to the home country. 
    Among the cultural dimensions, Individualism correlates the most with the DA index, with the correlation coefficient of 0.155. This means that higher the difference between the home and the destination country in terms of individualism, the higher a migrant's DA level. The same can be observed for masculinity: higher cultural differences result in higher DA. A contrasting picture is provided for the HA index: we see that it is significantly negatively correlated with individualism and masculinity. This means that the higher the difference between the home and the destination country in terms of individualism and masculinity, the less a migrant remains attached to their home country.

    Among the other variables, in general absolute correlations are rather low. The distance appears to be significantly related to both of our DA and HA indices: the further the destination country is to the country of origin, the higher the DA level and the lower the HA level. Also, the correlation between \textit{contig} and HA indicates that immigrants in destination countries where they share the border with their country of origin have higher HA levels. This makes sense since having the home country close means more possibilities to go back home frequently resulting in higher HA levels. 
    For the variables concerning language, the DA index is significantly positively correlated with all of them. The positive relationship between the DA index and the \textit{csl} highlights that the ease of communication is as important as having common native language or common official language for higher DA.

    \begin{table*}[t]
    \centering
    \resizebox{\textwidth}{!}{%
    \begin{tabular}{|l|l|l|l|l|l|l|l|l|l|l|l|}
    \hline
     & Power & Individualism & Masculinity & Uncertainty & Orientation & Indulgence & contig & comlang\_off & distcap & csl & cnl \\ \hline
    DA & -0.032 & 0.215** & 0.281*** & 0.164* & -0.09 & 0.028 & 0.427*** & 0.121 & 0.194* & 0.138 & 0.257** \\ \hline

    HA & 0.126 & -0.301*** & -0.164* & -0.094 & -0.03 & -0.159 & 0.343*** & 0.215* & 0.061 & 0.257** & 0.385*** \\ \hline
    \end{tabular}%
    }
    \caption{Correlation table for HA \& DA and Hofstede's cultural dimension scores. Correlation with HA is computed after grouping migrants by nationality, while correlation with DA is computed after grouping by residence. Significance levels are marked with *** p-value $<$0.01, ** p-value $<$0.05, * p-value $<$0.1. }
    \label{tab:hof_group}
    \end{table*}
    
    As already noted, absolute correlation values above are quite low, albeit significant. This is most probably due to individual differences within groups of migrants with the same nationality and residence, which decrease the correlations. To account for this, we repeat the correlation analysis, after grouping the migrants. Specifically, we group the migrants by nationality to compute correlations with HA levels, and by residence to compute correlations to DA levels. This allows us to have, for each home and destination country, an average HA and DA level, computed over a group of migrants. 
    
    The correlations obtained are shown in table \ref{tab:hof_group}. We note that grouping increased the correlations observed, confirming that the previous low correlations were due to individual variability, which averages out when grouping. Among the cultural dimensions, Individualism and Masculinity remain the most correlated, with the sign of the relation from the individual analysis confirmed. We observe an additional positive relation between Uncertainty and DA: the higher the difference in uncertainty the more the migrants are attached to the destination country. Regarding the other variables, grouping the migrants also increased the correlations significantly, and now the picture is clearer. It appears that the closer the home and destination countries are in terms of language, the higher the DA and HA levels. This confirms what we saw earlier, language is not important only for DA, but also for HA. In this case, having a common spoken/national/official language with the destination country allows migrants to maintain stronger links also with their home country. The same applies when home and destination countries share borders: both HA and DA are higher. In terms of geographical distance between capitals, we observe a weaker positive correlation with DA significant at 5\% level. This would indicate that the larger the distance among capitals, the more migrants become attached to the destination. While this could appear to contradict the results obtain with the contig variable, this is not necessary the case: it may be very well possible that neighbouring countries have large distances among capitals (especially non European countries) and vice-versa non neighbouring countries have small distances between capitals.

\section{Discussion}
\label{sec:discussion}

    In this work, we have developed a novel method to study cultural integration patterns of migrants through Twitter.
    Different from the existing literature, here we introduced hashtags from Twitter as a proxy for links to cultural traits of the country of origin or of the country of destination, which we call \textit{home attachment (HA)} and \textit{destination attachment (DA)}, respectively. 
    The HA and DA were defined by taking the proportions of country-specific hashtags that either belongs to the country of residence (DA) or the country of nationality (HA). 
    The null model analysis performed to validate the indices showed a significant difference between the actual indices and the null model indices, confirming the validity of our approach.
    The comparison between the indices and other related variables allowed us to discover interesting relations. 
    First, the proficiency of the language of the host country corresponds to higher DA level, as does having a common native language with the destination country Interestingly, common languages also correlate with large HA levels, which is a less explored result. 
    Second, we saw that in general, sharing borders increases both the DA and HA level. At the same time, the further the destination country, the higher the DA level.
    Through the comparison with Hofstede's cultural dimensions, we found that the higher the differences between the origin and destination countries in terms of individualism, masculinity and uncertainty, the higher the DA level is. These relationships are found to be the opposite with the HA index.

    It is important to mention that detecting causality in the results above may be difficult. For example, the proficiency of the language of the host country could facilitate higher DA levels. But this relationship could also be true the other way around. Although the causality issue cannot be disentangled in our analysis, we believe that through this work, we were able to shed light to important relationships between several important elements of culture and migrants' attachments to the host and home countries, thus highlighting  different cultural integration processes.

    Having employed social big data for our analysis came with several advantages. We were able to observe real-world social behaviour in an uncontrolled environment, avoiding the risk of having evasive answers, or/and misinterpretation of questions when completing a survey. 
    In addition, unlike surveys which often are incomparable across countries, we were able to conduct a cross-country study of integration of international migrants. It is important to note, however, that employing big data also has its drawbacks. Although we began with a total of about 60 million users, we ended up working with only 3,226 identified international migrants mainly due to the lack of geo-tagged tweets.  This shows that such a study requires very extensive resources to be completed.
    This analysis also suffers from sampling bias. 
    The Twitter population is different from the real one, hence not all the demographic groups are covered in the analysis. 
    Importantly, privacy and ethical aspects are often raised when using big data that contain personal information, even if the information was made public by the individuals themselves. This becomes particularly important when dealing with specific populations of minorities such as migrants. In this work, neither personal information nor migration status of individuals has been released at any stage of the analysis. The data was securely stored and accessed. All results are aggregated at national level and presented in such a way that re-identification is not possible. In addition, we need to underline the fact that the findings of this paper cannot be generalised. They apply solely to a small sample of the population, and not to larger groups. The study has passed ethics approval before publication.

\bibliographystyle{splncs04}
\bibliography{biblio}

\end{document}